\documentclass{article}

\usepackage{microtype}
\usepackage{graphicx}
\usepackage{subfigure}
\usepackage{booktabs} 
\usepackage{pgf}
\usepackage{graphicx}
\graphicspath{{images/}}
\usepackage{bm}
\usepackage{blindtext}

\usepackage{subfiles}
\usepackage{tikz}
\usetikzlibrary{positioning,arrows}
\usetikzlibrary{shapes.geometric}
\usepackage{import}
  
\usepackage{hyperref}

\usepackage[accepted]{icml2025}

\usepackage{amsmath}
\usepackage{amssymb}
\usepackage{mathtools}
\usepackage{amsthm}

\usepackage[capitalize,noabbrev]{cleveref}

\newtheorem{assum}{Assumption}
\newtheorem{definition}{Definition}[] 
\newcommand{\KL}{\textnormal{KL}}

\newcommand{\Pc}{{\cal P}}

\newcommand{\Ac}{\mathcal{A}}

\newcommand{\omegaV}{\boldsymbol{\omega}}
\newcommand{\mtbE}{\mathbb{E}}

\newcommand{\TlP}{\mu}
\newcommand{\KLinfL}[0]{\operatorname{KL_{inf}^L}}
\newcommand{\KLinfU}[0]{\operatorname{KL_{inf}^U}}
\definecolor{Gans}{rgb}{0.3, 0.7, 0.3}

\usepackage[textsize=tiny]{todonotes}

\icmltitlerunning{Identifying the Best Transition Law}

\begin{document}

\twocolumn[
\icmltitle{Identifying the Best Transition Law}
  \begin{center}
 {\textbf{Mehrasa Ahmadipour}, \quad \textbf{élise Crepon}, \quad \textbf{Aurélien Garivier} \\  
    UMPA, ENS de Lyon, Lyon, France }
         
  \end{center} 



\icmlsetsymbol{equal}{*}



\icmlcorrespondingauthor{Mehrasa Ahmadipour}{Mehrasa.ahmadipour@ens-lyon.fr}

\icmlkeywords{Machine Learning, ICML}

\vskip 0.3in
]



\section{Abstract}
Motivated by recursive learning in Markov Decision Processes, this paper studies best-arm identification in bandit problems where each arm's reward is drawn from a multinomial distribution with a known support.
We compare the performance { reached by strategies including notably LUCB without and with  use of this knowledge. }
In the first case, we use classical non-parametric approaches for the confidence intervals. In the second case, where a probability distribution is to be estimated, we first use classical deviation bounds (Hoeffding and Bernstein) on each dimension independently, and then the Empirical Likelihood method (EL-LUCB) on the joint probability vector. 
The effectiveness of these methods is demonstrated through simulations on scenarios with varying levels of structural complexity.
\section{Introduction}
The importance of interactions between entities such as human-computer interfaces, complex decision-making in autonomous systems like self-driving cars \cite{car}, or dynamic difficulty adjustment (DDA) in online gaming \cite{DDAsurvey} has fostered the development of Reinforcement Learning (RL) as a model for dynamical systems where responsible agents aim for optimal decisions in an uncertain environment. Markov Decision Processes (MDP) proved able to capture many interesting features of these scenarios, while providing a rich toolbox of computationally efficient and mathematically founded algorithms\cite{puterman1994mdp,bertsekas2005dp,sutton2018reinforcement,RL-survey}.
 
An MDP is defined as a tuple consisting of the state space $\mathcal{S}$, the action set $\mathcal{A}$,  the transition probability  kernel $\mathcal{P}$, and   the reward function $\mathcal{R}$. At time $t$, a (deterministic) policy is a mapping $\pi_t: \mathcal{S}\to\mathcal{A}$ that specifies which action to choose according to the current state. 
To determine the optimal policy in an MDP, \emph{Bellman equations} provide a recursive formulation for the value function  $V_t(s)$, the expected cumulative reward starting from state $s$ at time $t$. In the case of  a finite horizon $T$, the optimal policy $\pi^*$ satisfies:
\begin{equation}
V^{\pi^*}_t(s) = \max_{\pi \in \Pi } \left[ \mathcal{R}(s, a) + \sum_{s' \in \mathcal{S}} \mathcal{P}(s' \mid s, a) V^{\pi^*}_{t+1}(s') \right]. \label{eq:bellman_optimal}
\end{equation}

When the transition probabilities are fully known in the problem, the agent can compute an optimal policy without interacting with the environment -- a process known as {\it Planning }(see \cite{moerland2023model}). Otherwise, the process of {\it learning} requires the estimation of expected future returns.
{
This paper focuses on a particular learning sub-task: the choice of the policy at time $t$ with known value function $V_{t+1}$\footnote{the use of this sub-task in a backward induction for a step-by-step solution to the dynamic programming formulation is left for future work.}. The player, in state $s$,  can transition to one of $d$ possible next states $s'$, each associated with a certain value $V_{t+1}(s')$.  
She chooses an action $a\in\Ac$, and transitions to a next state $S^{\prime}$ according to the transition probability vector $P_a$ (see Figure~\ref{fig:systemmodel}).
Each destination state has an associated expected value, and the goal is to find the best action that maximizes this expected value.


\begin{figure}[h]
\centering

\begin{tikzpicture}[auto, node distance=8mm, >=latex, font=\small]
    \tikzstyle{round} = [thick, draw=black, circle]
    
    \node[round] (s0) {$s$};
    
    \node[round, below left=10mm and 20mm of s0] (s1) {$s'_1$};
    \node[below=10mm of s0] (s2) {$\cdots$};
    \node[round, below right=10mm and 20mm of s0] (s3) {$s'_d$};

    \draw[->, black, thick,line width=0.6mm] (s0) -- (s1);

    \draw[->, blue, thick, line width=0.9mm] (s0) to[out=140, in=90] (s1);

    \draw[->, red, thick,line width=0.3mm] (s0) to[out=250, in=-20] (s1);

    \draw[->, blue, thick,line width=0.3mm] (s0) to[out=40, in=90] (s3);
    \draw[->, black, thick,line width=0.6mm] (s0) -- (s3);
    \draw[->, red, thick,line width=0.9mm] (s0) to[out=-80, in=-160] (s3);
\end{tikzpicture}

\caption{A learner at  $s$  selects a color and transitions to  $S^{\prime}$. Each color represents a probability vector, meaning that the likelihood of arriving at a destination varies depending on the chosen color.}
\label{fig:systemmodel}
\end{figure}
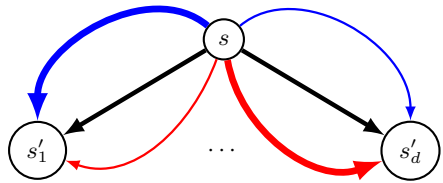

This decision-making problem is very reminiscent of a Multi-Armed Bandit (MAB) problem, as the outcome of each action  $a$  depends only on the current state  $s$  and transition probabilities  $P_a$, independently of past decisions. The considered task is usually called Best Arm Identification (BAI): identify the arm yielding the highest expected value with as few samples as possible. 

But the considered bandit problem has a specific feature: the support of the arms, $\{V_{s'}: s'\in\mathcal{S}\}$, is known for all arms. This feature can be used or not by the agent: in the \emph{non-structured setting}, the agent ignores these values and simply adopts a classical bandit strategy. She can then rely on established algorithms such as LUCB \cite{Kalyanakrishnan2012LUCB} and Track\&Stop ~\cite{Garivier16}.
In the \emph{structured setting}, each arm is modeled as multinomial distribution on a known support -- see also~\cite{agrawal2020optimal}. For this case, we propose Structured-LUCB, a modified version of the LUCB algorithm, and an Empirical Likelihood  approach (EL-LUCB) inspired by \cite{SaraF2010}. The goal of this paper is to investigate in which way the structured approach outperforms the non-structured one in specific scenarios on the support of the distribution, while in others, the non-structured method can perform better.
}
\section{State-of-the-art and connections}

In this work, we focus on the fixed-confidence PAC (Probably Approximately Correct) setting for MAB problems. 
While the fixed-budget setting remains somewhat mysterious (see e.g., \cite{bubeck2012regret}), the fixed-confidence setting is better understood since its sample complexity was identified in~\cite{Garivier16} thanks to change-of-measure techniques.
Different structures for bandit arms have been considered: multi-modal \cite{saber2024bandits}, linear \cite{abbasi2011improved}, contextual\cite{lu2010contextual}, kernel-based models\cite{Vakili2024}, etc. However, these models do not address bandit problems with multinomial reward distributions. We propose  novel adaptations of LUCB, {\it —Structured-LUCB} using Bernstein and Hoeffding inequalities— and an {\it EL-LUCB} algorithm.

In the Structured-LUCB, we use  empirical Bernstein bounds which have been investigated in UGapE \cite{gabillon2012best} and in the context of Racing algorithms \cite{Minh2008Bernstein,Heidrich2009Bernstein}, where Bernstein-based methods were used to design efficient stopping times. Bernstein-based concentration is also used in regret minimization in \cite{audibert2010best}. The empirical likelihood method seems particularly well suited for multinomial distributions. In the EL-LUCB algorithm, we employ Kullback–Leibler (KL)-divergence-based confidence regions with LUCB, inspired by the regret minimization algorithms of \cite{SaraF2010},~\cite{kaufmann2013KLLUCB} or~\cite{cappe2013kullback}. These structured methods aim to improve performance by incorporating additional information about the underlying probability vector and its constraints on a simplex.

Our study compares these two methodologies under various scenarios. We employ the Top-two (leader-challenger) sampling rule \cite{russo2016simple,jourdan2022top,you2023information}, which has shown robust performance in both Bayesian and frequentist settings, to guide our sampling strategies. Recent work by \cite{jourdan2022top} extends this method to bounded distributions, and \cite{you2023information} presents a further enhancement with theoretical guarantees.

By contrasting Structured and Non-Structured algorithms, we explore whether leveraging known structures can yield significant benefits in terms of sample efficiency and decision accuracy. To the best of our knowledge, no prior research has systematically examined shifting perspectives between the two approaches. Our contribution lies in this comparative analysis, providing insights into when and how structural assumptions can be beneficial.

The paper is organized as follows. We begin by formally explaining the model and our assumptions. We then develop the non-structured approach before the structured cases. We finally present the numerical experiments that we compare and discuss on different algorithms. 

\section{System Model}\label{sec:sys_model}
We consider $K$ multinomial distributions $P_1, \dots, P_K$. Each distribution has $d$ mutually exclusive outcomes, associated with values $V = [v_1,\dots,v_d] \in \mathbb{R}^d$. An outcome $v_i$ represents the value obtained at the next state $S^{\prime}$.
 We describe each distribution as a vector  $P_a = [p_{a,1}, \dots, p_{a,d}]$  lies on the simplex
$\Delta^d :=$ $ \Big\{ \theta \in \mathbb{R}^{d+1} \ \Big| \ \sum_{i=1}^{d+1} \theta_i = 1, \ \theta_i \geq 0 \text{ for all } i = 1, \dots, d \Big\}$,
 i.e., $p_{a,i} \geq 0$ and $\sum_{i=1}^d p_{a,i} = 1$.  
 
 At discrete time intervals $t = 1, 2, \dots$, the learner selects an action $A_t \in \mathcal{A}$ and receives an independent sample $Z_{A_t} = [Z_{A_t,1}, \cdots, Z_{A_t,d}]$ where we assume $Z_{A_t}$ is a one-hot vector indicating the next state $S^{\prime}$. Specifically, $Z_{A_t}$ is drawn such that $\mathbb{P}[Z_{A_t} = e_i] = p_{A_t,i}$, where $e_i$ is the $i$-th standard basis vector in $\mathbb{R}^d$. 
 Denoting by $p\cdot v$ the scalar product of two vectors $p$ and $v$, the expected value of the reward  $V$  under the probability vector  $P_a $ is expressed as $\mathbb{E}_{P_a}[V] = \sum_{i=1}^d p_{a,i} v_i = P_a \cdot v$. 
 Without loss of generality, we can assume the following order:
\begin{equation}\label{Assume:bestarm}
\mathbb{E}_{P_1}[V] > \mathbb{E}_{P_2}[V] > \cdots > \mathbb{E}_{P_K}[V].
\end{equation}
The learner empirically constructs $\hat{P}_a$ and attempts to find the action $a^*$ that maximizes the expected reward as soon as possible:
\begin{equation}\label{object1}
\max_{a \in \mathcal{A}} P_a \cdot V  \quad \text{s.t. } \text{Dist}(\hat{P}_a, P_a ) \leq \epsilon,
\end{equation}
where $\text{Dist}(\cdot,\cdot)$ quantifies the “distance” between the estimated transition probabilities $\hat{P}_a$ and the optimistic transition probabilities $P_a$. We define this distance more precisely later, first using an $L$-norm, then using the $\KL$-divergence.

We operate in the \textit{fixed-confidence} regime, where a maximal risk parameter $\delta \in (0,1)$ is fixed.
\begin{definition}
A strategy is called $\delta$-PAC (Probably Approximately Correct) if, for every tuple of distributions $\mathcal{P} = (P_1, \dots, P_K)$, it satisfies $\mathbb{P}_{\mathcal{P}}[\tau < \infty] = 1$ and $\mathbb{P}_{\mathcal{P}}[\hat{a}^* \neq a^*] \leq \delta$.

\end{definition}
 	
	\begin{definition}[Sample Complexity]
Given a bandit model with $K$ arms, the fixed-confidence sample complexity $\kappa$ is defined as the minimum expected number of samples needed by a $\delta$-PAC algorithm:
\begin{equation}
\kappa := \inf_{\text{PAC algorithms}} \limsup_{\delta \to 0} \frac{\mathbb{E}[\tau]}{\log \frac{1}{\delta}}.
\end{equation}
\end{definition} 
 To achieve the goal of identifying the best arm, the learner must employ a strategy denoted by the triple \(\Big( (A_t), \tau, \hat{a}^* \Big)\), which includes a sampling rule \(A_t\) determining the chosen arm based on past actions and rewards, a stopping rule \(\tau\), and a recommendation rule \(\hat{a}^*\) that suggests the best action at termination.
 The learner’s strategy is crucial for efficiently identifying the best arm, requiring a careful balance between exploration and exploitation, while considering the observed outcomes to make informed decisions about arm selection and termination of the sampling process.

For simplicity, we assume that the rewards are bounded:
\begin{assum}\label{assum_V}
Assume that  $v_i \in [0,1]$  for all  $i \in \{1, \dots, d\}$. 
\end{assum}
\vspace{-0.3cm}
Since the deviations of any $[0,1]$ random variables from its expectation are bounded by those of a Bernoulli distribution with the same mean, this assumption allows us to model the expected reward using a Bernoulli distribution with parameter  $\mu_a := \mathbb{E}_{P_a}[V]$.
Our initial approach to the problem employs the known BAI methods within the MAB for Single Parameter Exponential Family(SPEF) distributions. 

\section{Non-Structured Approach}\label{system_SoArt}
We assume that rewards are i.i.d. and follow a Bernoulli distribution with parameter $\mu_a := \mathbb{E}_{P_a}[V]$ under Assumption~\ref{assum_V}. This setting leads to a MAB problem with distributions belong to the SPEF as described in \cite{Garivier16}. 
The concept of distinguishability is employed to characterize the lower bounds for the sample complexity in \cite{Garivier16}. This notion is quantified using the KL-divergence, denoted as  $\KL(x, y) := x \log(\frac{x}{y}) + (1 - x) \log(\frac{(1 - x)}{(1 - y)})$.
Denote by $\mathcal{S}$ a set of SPEF bandit models such that each bandit model $\bm\mu = (\mu_1, \dots, \mu_K)$ in $\mathcal{S}$ has a unique optimal arm $a^*(\bm\mu)$.
For each $\bm\mu \in \mathcal{S}$, there exists an arm $a^*({\bm\mu})$ s.t. 
They use the notion of the alternative set
\[
\operatorname{Alt}(\bm\mu)
\;:=\;
\{\Lambda \in \mathcal{S} \;:\; a^*({\Lambda}) \,\neq\, a^*(\bm\mu)\},
\]
which is the set of problems $\Lambda$ for which the optimal arm $a^*(\lambda)$ differs from the optimal arm $a^*(\mu)$ of the reference distribution $\bm\mu$. They characterize a lower bound for any $\delta$-PAC strategy and any bandit model $\bm\mu \in \mathcal{S}$ under a given risk level $\delta \in (0,1)$:
		\begin{equation}
			\mtbE [\tau_{\delta}]\geq T^*(\bm\mu) \KL (\delta, 1-\delta),
   \label{Garivier:Lower}
		\end{equation}
		where 
		\begin{equation}
			T^* (\bm\mu )^{-1} := 	\sup_{\omegaV \in \Sigma_K} 
			\inf_{\substack{Q \in \textnormal{Alt}(\bm\mu) }}  \quad \sum_{a=1}^{K}\omega_a \KL(\TlP_a, Q_a).
			\label{Topt}
		\end{equation}
Here, the set of proportions for pulling arms is defined as
$\Sigma_K
\;=\;
\{\omega \in \mathbb{R}_+^K \;:\; \sum_{i=1}^K \omega_i \;=\; 1\}$.

The assumption of SPEF  on bandits' distributions allows the inner minimization of \eqref{Topt} to be solved and to obtain an explicit formulation for the optimal weights \(\omega\). The same authors introduced an asymptotic optimal algorithm matching this lower bound, called Track\&Stop (T\&S). But its computational complexity often motivates the use of more practical alternatives such as the LUCB (Lower and Upper Confidence Bound) algorithm.

\subsection{Non-structured LUCB}
\label{sec:lucb_nostruct}
The LUCB algorithm is a standard approach for the PAC problem in stochastic multi-armed bandits, specifically for BAI. The main idea is to construct and iteratively refine upper and lower confidence bounds around each arm’s empirical mean \cite{Kalyanakrishnan2012LUCB} until they are separated.

At each round \(t\), for each arm \(a\), we compute a confidence bonus (CB) \(\beta^{\text{No-St}}_a(n,t)\). Over time, \(\beta^{\text{No-St}}_a(n,t)\) shrinks according to a concentration inequality (e.g., Hoeffding or Bernstein).  Let \(\hat{\mu}_a(t):=\hat{P}_a(t)\cdot V\) be the empirical estimate of arm $a$'s expectation at time \(t\). For the current best arm \(a\) and the current second best arm \(b\), we construct lower confidence bound and upper confidence bound respectively. The algorithm stops when
\begin{equation}
  \hat{\mu}_a(t)-\hat{\mu}_b(t) 
    -\bigl[\beta^{\text{No-St}}_a(n,t)+ \beta^{\text{No-St}}_b(n,t)\bigr]
\geq \epsilon,
\label{stp:lucb}
\end{equation}
where 
$
    \beta_i^{\text{No-St}}\bigl(n_i^t,\,t\bigr)
    \;=\;
    \sqrt{
        \frac{
            \log\Bigl(\frac{2Kt}{\delta}\Bigr)
        }{
            2\,n_{i}(t)
        }
    }$ is derived from  Hoeffding’s inequality.
We call this approach Non-structured because the algorithm ignores the vector $V$. 
\section{
Structured System Model  }
\label{sec:str1}



We now consider the model introduced in Section~\ref{sec:sys_model} as a  $K$-action bandit problem where rewards are drawn i.i.d. from \textit{multinomial} distributions  $P_a, a\in \Ac$  and  $V = [v_1, v_2, \ldots, v_d]$ is the support. 
We rely here directly on estimates of all $d$ components of the probability vector $P_a$. We construct the empirical distribution as a vector 
$\hat{P}_{A_t} = \frac{1}{n_{A_t}} \sum_{i=1}^{n_{A_t}} \delta_{Z_i}$ where each component is given by $\hat{P}_{A_t, i} = \frac{1}{n_{A_t}} \sum_{k=1}^{n_{A_t}} \mathbb{I}(Z_{A_t,k} = e_i)$ with  $\mathbb{I}$  being the indicator function.
We keep the Assumption~\ref{assum_V} to facilitate the comparison of previously non-structured approach with the proposed structured cases.
We address a more general case involving bounded support probabilities and heavy-tail distributions, as introduced and analyzed in \cite{agrawal2020optimal}.  While SPEF distributions in \cite{Garivier16} allow inner minimization of \eqref{Topt} in Euclidean space, introducing a probability vector confines the problem to the simplex. Agrawal et al. \cite{agrawal2020optimal} address this by using functions \(\KLinfL\) and \(\KLinfU\), which measure the minimal $\KL$-divergence required to distinguish a distribution $\eta$ from alternatives with means below or above a threshold $x$. Specifically, for distributions $\kappa_1,\kappa_2$ over a finite set with mean $m(\kappa)$, define
\[
\KLinfU(\eta,x) :=
\min_{\substack{\kappa \in \mathcal{L}\\ m(\kappa)\,\ge\,x}}
\KL(\eta,\kappa),
\]
with a similar definition for \(\KLinfL(\eta,x)\). 
Inspired by \cite{Honda2010AnAO} in regret minimization setup, Agrawal et al. propose a  Lagrangian dual problem to overcome the technical challenge of lower bounding sample complexity in this simplex-based model and proved \eqref{Garivier:Lower} with new definition of inner minimization of $T^* (\Pc )$ using \(\KLinfL\) and \(\KLinfU\). In \cite{agrawal2020optimal}, a modified version of T\&S  is proposed.
	 
{ 
The computational complexity issue is further exacerbated in the modified T\&S algorithm, where the complexity increases due to the need to solve an optimization problem involving two $\KL_{\text{inf}}$ terms. 
Additionally, incorporating the effect of $V$ when transitioning from Track-and-Stop to modified Track-and-Stop is not straightforward. In the latter, the support vector independently influences the Lagrangian problem, making it difficult to unify and clearly reflect the impact of $V$. These challenges motivate us to explore the Structured-LUCB algorithm practically, where the influence of $V$ on the algorithm becomes more transparent  to analyze.}

\subsection{Structured-LUCB } 

The Structured-LUCB algorithm constructs confidence bounds for each $d$ component of the probability vectors and combines them to compute confidence intervals for expected rewards. Each  $p_{a,i}$  is estimated independently. The algorithm applies larger thresholds in decision-making for components  with $v_i$  that are more likely to occur, prioritizing areas of higher uncertainty. Algorithm~\ref{alg:structured_lucb_prob} provides the schema for the Structured-LUCB algorithm.
At time $t$, let $\hat{p}_{k,i}(t)$  denote the empirical estimate for the probability of arm $k$ producing outcome $v_i$, and let $\beta^{\text{St}}_{k,i}(n,t)$  quantify the uncertainty of this estimate based on  $n$  samples.
Following the LUCB framework, we require a lower bound for the best arm  $a$  and an upper bound for the current second-best arm  $b$  on the  $i$-th outcome:
\begin{align}
&p_{a,i} \geq \hat{p}_{a,i}(t) - \beta^{\text{St}}_{a,i}(n,t), \\
&p_{b,i} \leq \hat{p}_{b,i}(t) + \beta^{\text{St}}_{b,i}(n,t),
\end{align}
where the CB  $\beta^{\text{St}}_{k,i}(n,t)$  is derived using either Hoeffding’s inequality:
\begin{align}\label{beta:CB:HF}
&\beta^{\text{Str-H}}_{k,i}(n,t) = 
\sqrt{ \frac{ \log( \frac{2 d K t}{\delta} ) }{ 2 n_{k}(t) } }, 
\end{align}
or Bernstein’s inequality, which incorporates the variance for tighter bounds:
\begin{align}
\beta^{\text{ST-B}}_{k,i} (t) = 
\sqrt{ 2 \hat{\sigma}_{k,i,t}^2 }
\sqrt{ \frac{\ln\left( \frac{2 d K  t}{\delta} \right)}{2n_k} } + \frac{ \ln\left( \frac{2 d K t }{\delta} \right) }{3n_k },
\label{beta_BErn}
\end{align}
where $\hat{\sigma}_{k,i,t}^2$  denotes the variance at time  $t$ , calculated as  $\hat{\sigma}_{k,i,t}^2 = \hat{p}_{k,i}(t)(1 - \hat{p}_{k,i}(t))$, providing confidence bonuses (CBs) that depend more closely on the observed variance.

The lower and upper confidence bounds for the expected rewards of arms  $a$  and  $b$  are given by :
\begin{align}\label{eq:LUCB_construction}
&\text{LCB}^{\text{str}}_a(t) = \sum_{i=1}^d \left( \hat{p}_{a,i}(t) - \beta^{\text{St}}_{a,i}(n,t) \right) v_i, 
\\
&\text{UCB}^{\text{str}}_b(t) = \sum_{i=1}^d \left( \hat{p}_{b,i}(t) + \beta^{\text{St}}_{b,i}(n,t) \right) v_i.
\end{align}
The algorithm  stops when  $\text{LCB}{^\text{str}}$  exceeds  $\text{UCB}{^\text{str}}$ , ensuring that the expected reward of arm  $a$  is sufficiently higher than that of arm  $b$  with high confidence.
Since the confidence bounds are uniform across all components of each probability vector, the stopping condition is simplified to:
\begin{align}
\label{STOPT:LUCBSTR}
&\left( \hat{P}_a(t) - \hat{P}_b(t) \right)\cdot V 
- \big(  \beta^{\text{St}}_{a,i}(n,t) + \beta^{\text{St}}_{b,i}(n,t)\big)\cdot |V| \geq \epsilon,
\end{align}
where $|V|$ is $L_1$-norm of vector $V$. 
\begin{algorithm}[t]
\caption{Structured-LUCB Algorithm with Leader-Challenger Sampling Strategy}
\label{alg:structured_lucb_prob}
\begin{algorithmic}[1]
    \STATE \textbf{Input:}  $V$ ,  $\delta$,  $\alpha \in [0,1]$
    \STATE \textbf{Output:} Optimal arm $a^*$
    
    \STATE \textbf{Initialization:}
  $\forall a \in \Ac$,   $n_a \gets 1$, $\forall a \in [K]$, observing the reward  $Z_a$ and  $\hat{P}_{a} \gets Z_a$
  
    \WHILE{$\text{LCB}_l^{\text{str}}-\text{UCB}^{\text{str}}_c(t)< \epsilon $}
        \FOR{$a \in [K]$}
            \STATE Choose $\beta=\min ( \beta^{\text{Str-H}}_{k,i}, \beta^{\text{ST-B}}_{k,i})$ based  on \eqref{beta:CB:HF}- \eqref{beta_BErn}
            \STATE Construct $(\text{LCB}_a^{\text{str}},\text{UCB}_a^{\text{str}})$, $\forall a\in\Ac$
        \ENDFOR
        
        \STATE $l \gets \arg\max_a \text{LCB}^{\text{str}}_a$
        \STATE $c \gets \arg\max_{a \neq a_{\text{leader}}} \text{UCB}^{\text{str}}_a$
        \STATE Assign $a_t \gets 
        \begin{cases} 
            l & \text{if } X = 1 \sim \text{Bernoulli}(\alpha), \\
            c & \text{otherwise.}
        \end{cases}$
        
        \STATE \textbf{Observe:} Reward $Z_{a_t}$ from pulling arm $a_t$
        \STATE \textbf{update the parameters}
    \ENDWHILE
\end{algorithmic}
\end{algorithm}
Next in this section, we look at a joint estimation of $d$ components of the probability vectors.
{
\subsection{Structured Model~2: EL-LUCB Method}\label{sub:saraFillipi2010}
The problem introduced in \eqref{object1} can be interpreted as a  maximization of  an unknown probability vector  $P_a$  in the direction of a known vector  $V$. While the Structured-LUCB algorithm considers this problem as an estimation of all component independently, we suggest here to think of a joint estimation within a {\it $\KL$-ball}, that reminiscent of  the approach in \cite{SaraF2010} for a regret minimization problem. 
In previous Structured-LUCB,  the $\text{Dist}(\cdot,\cdot)$ was defined as an $L$-norm while here we use $\text{Dist}(\hat{P}_a,P_a):=\KL(\hat{P}_a, P_a ) \leq \epsilon$.

The primary modification, compared to Algorithm~\ref{alg:structured_lucb_prob}, appears on the seventh line, where the construction of the $\text{LCB}$s and $\text{UCB}$s is specified.
At each iteration, given the updated empirical distributions of the leader arm $\hat{P}_a$ and the challenger arm $\hat{P}_a$, we  apply Algorithm~2 from \cite{SaraF2010} to obtain respectively  a lower bound on $\hat{P}_a(t) \cdot V$ and an upper bound on $\hat{P}_b(t) \cdot V$.

At each iteration of the loop, the updated $\KL-$ball, or more specifically the $\KL$-ellipses, around the estimates shrinks until there is no overlap, indicating that the estimates have reached a desired precision. The Algorithm~2As of \cite{SaraF2010} is solved using a Lagrangian multiplier approach.

\section{Experiments and Discussions}

In this section, we explore our proposed algorithms for different cases and explain the different results obtained.
We focus on three algorithms: Non-structured LUCB (or simply LUCB with Assumption~\ref{assum_V}), Structured-LUCB presented in Alg~\ref{alg:structured_lucb_prob} and the EL-LUCB algorithm explained in Subsection~\ref{sub:saraFillipi2010}.

The results are averaged on 100 trials. 
The confidence parameter is set to $ \delta = 0.05$,  the sampling probabilities of leader-challenger are initialized  at $[0.5, 0.5]$.
The reward of each arm is drawn according to a row of the matrix  $P$. Its columns contain the probabilities of each outcome of $V$. We evaluate their performance on different support vectors $V$.

\subsection{ Structured-LUCB vs. Non-structured-LUCB}
In the situations where the outcome probabilities are highly concentrated on a single outcome of $V$ for each arm, the structured algorithm does not offer significant advantages over the non-structured algorithm. Both algorithms will perform similarly and  non-structured algorithm can quickly and accurately estimate the expected rewards based on observed averages with less complex implementation effort.
We hence consider the following cases to determine the suitability of each algorithm under varying conditions: 
$$
\begin{array}{lcl}
P^{test1} =
\begin{bmatrix}
0.5 & 0.3 & 0.2 \\
0.4 & 0.3 & 0.3 \\
0.3 & 0.2 & 0.5 
\end{bmatrix},
& \quad &
\begin{aligned}
V^{\text{test1}} &= [0.5, 0.1, 0].
\end{aligned}
\end{array}
$$
\begin{figure}[b!]
    \centering
    \begin{subfigure}{}
        \includegraphics[width=0.48\textwidth]{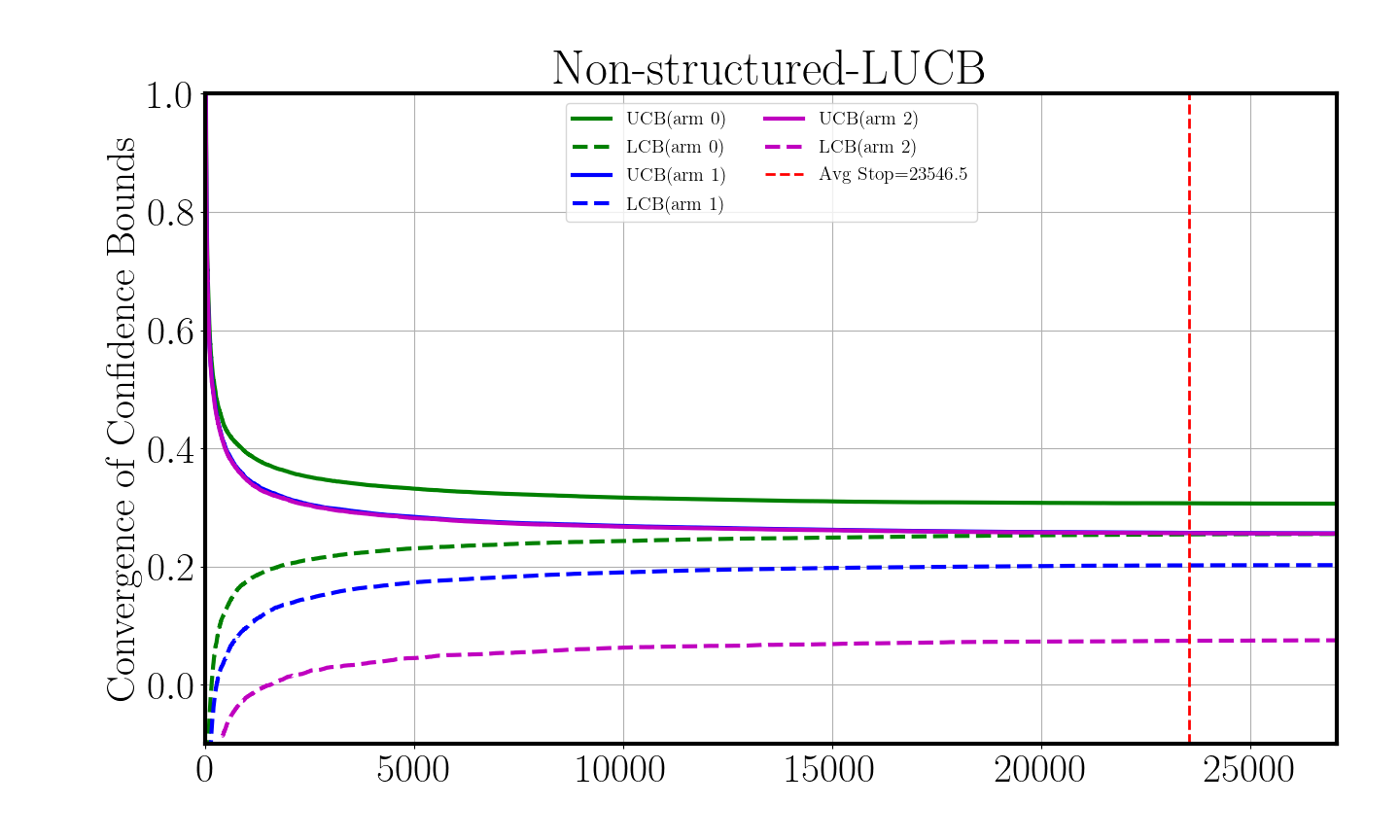}
    \end{subfigure}
    \begin{subfigure}{}
\includegraphics[width=0.5\textwidth]{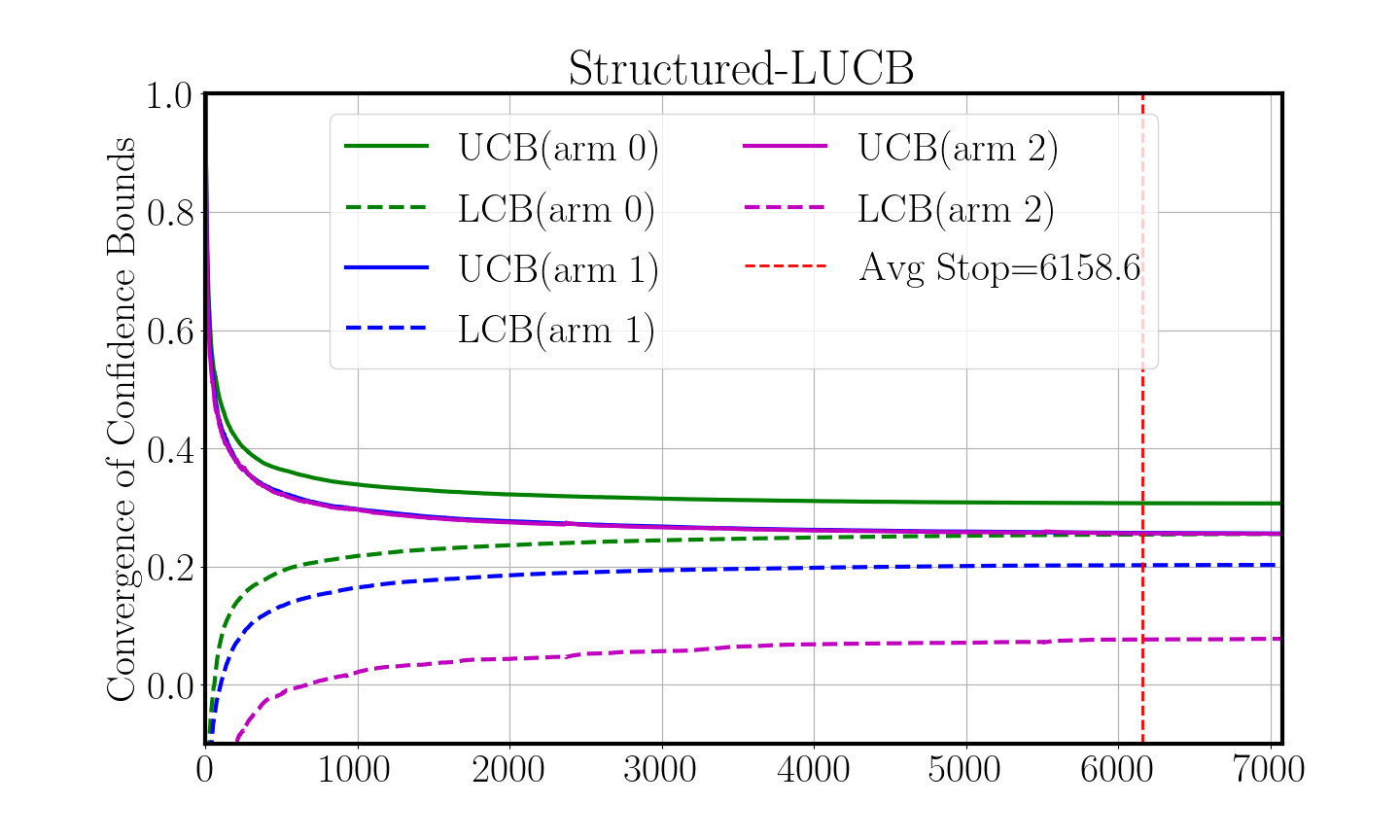}
    \end{subfigure}
    \caption{Comparing the stopping times of two algorithms  on $V^{\text{test1}}$}
     \label{str-wins}
    \end{figure}
Figure~\ref{str-wins} shows that for $V^{\text{test}1}$, the Structured-LUCB  algorithm significantly outperforms Non-structured-LUCB.

\begin{figure}[b!]
    \centering
    \begin{subfigure}{}
        \includegraphics[width=0.49\textwidth]{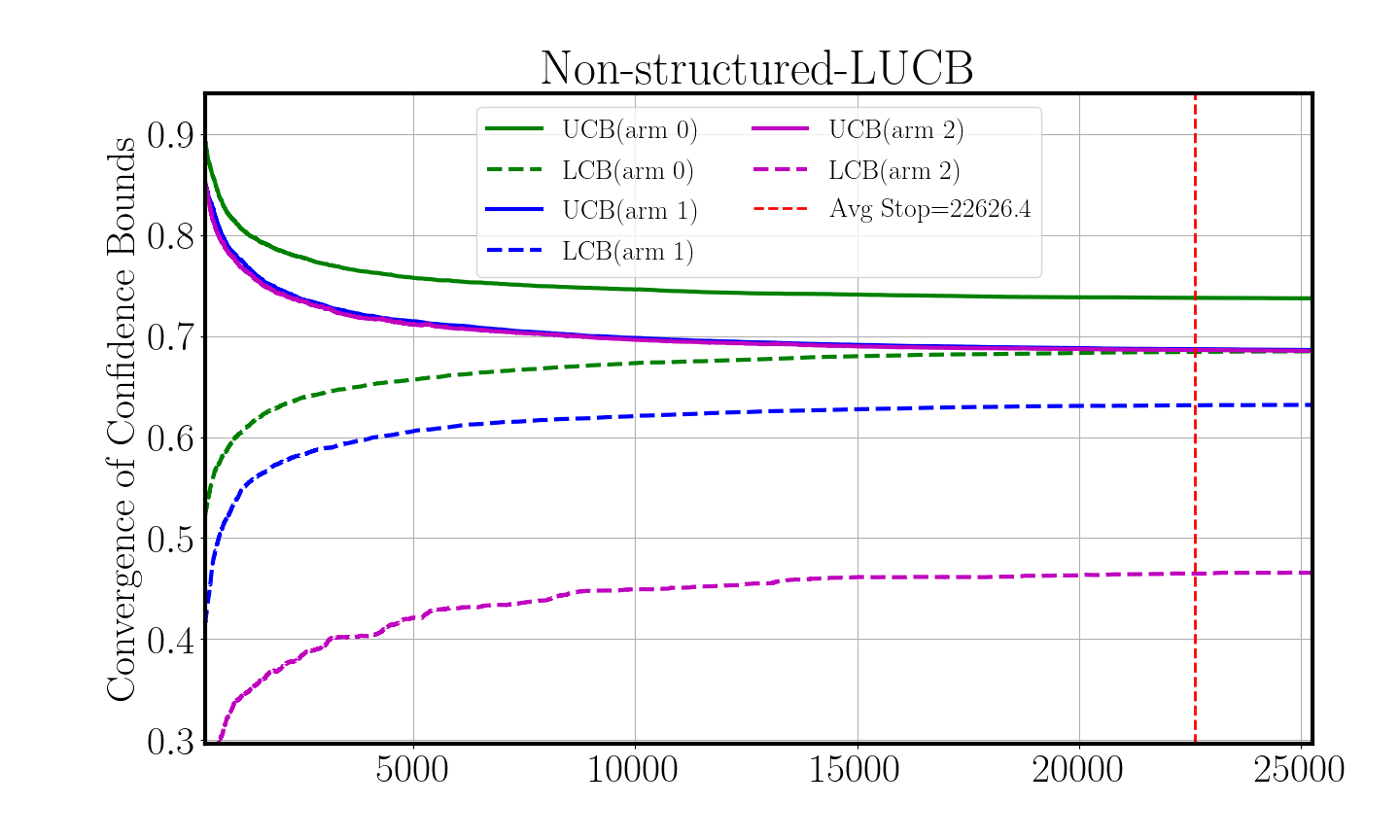}
    \end{subfigure}
    \begin{subfigure}{}
\includegraphics[width=0.48\textwidth]{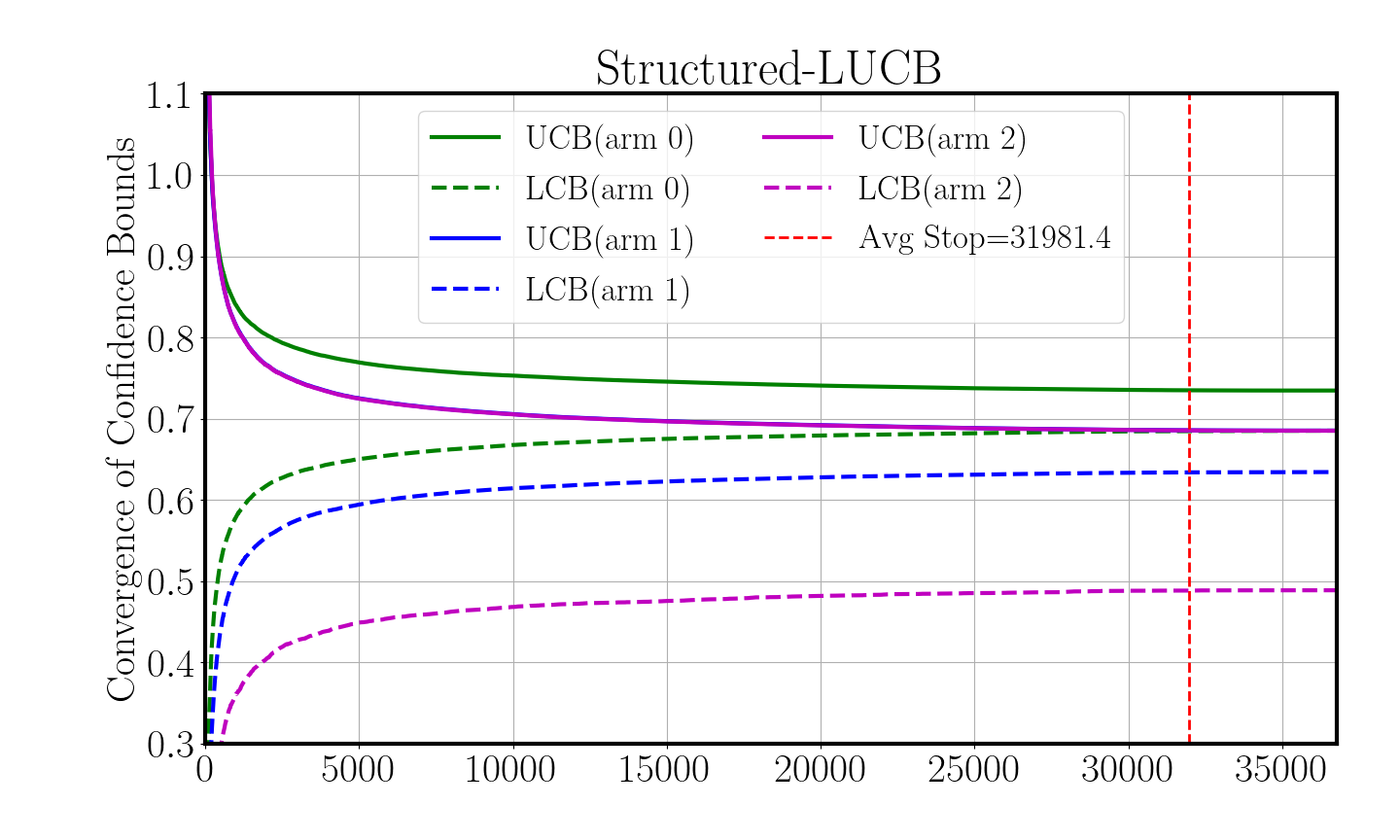}
    \end{subfigure}
    
    \caption{Comparing the stopping times of two algorithms on $V^{\text{test2}}$.}
\label{Nonstr-wins}
\end{figure}
For the same $P$, we change the support to  $V_{\text{test2}} = [0.9, 0.6, 0.4]$, as shown in Figure~\ref{Nonstr-wins}, Structured-LUCB demonstrates less efficient performance compared to Non-structured-LUCB.

The reasoning lies in how the stopping time and the effect of  $V$  are introduced in the algorithms. In Non-structured-LUCB, the summation  of CBs for Bernoulli distributions in 
\eqref{stp:lucb}
is not  effected by  $V$. However, in the stopping condition of the Structured-LUCB algorithm \eqref{STOPT:LUCBSTR},  $V$  plays a significant role. While the Hoeffding’s CB is scaled directly by $V$, the Bernstein’s CB is more affected by the features of $V$. Here, we provide the summation of Bernstein’s CBs for two arms in terms of Hoeffding’s CBs:
\begin{align}
  & \Delta^{\text{St-B}}
:=
\Big(
\frac{ \ln\left( \frac{2 d K t }{\delta} \right) }{3n^t_a }
 +\frac{ \ln\left( \frac{2 d K t }{\delta} \right) }{3n^t_b }\Big).(\sum_{i=1}^d v_i)
\nonumber\\
&+
\hspace{0cm}\sqrt{2 \sum_{i=1}^d \left(  \sigma_{a,i} \beta^{\text{ST-Hf}}_{a,i} + \sigma_{b,i} \beta^{\text{ST-Hf}}_{b,i} \right)^2 } \cdot ||V||.
\end{align}

In the structured case, the vector  $V$  is integrated into the stopping condition, making the algorithm sensitive to both individual outcomes  $v_i$  and the $l_1$-norm of $V$. This sensitivity is particularly evident when using Hoeffding’s bound and becomes more nuanced with Bernstein’s bound. 
By comparing the structured CB  to the non-structured CB, we can draw the following insights. When  $|V| \leq 1$, the Structured-LUCB stops earlier because its CBs shrink more rapidly. This accelerated reduction is due to the CB being scaled by  $|V|$. In contrast, when  $|V|\geq 1$ the Non-structured-LUCB algorithm becomes the preferred choice. However, if the number of states $d$ is very large ($d \gg 1$), it may offset the advantage of having $|V| \le 1$ in the structured scenario.

\textbf{The EL-LUCB Method}
LUCB-based algorithms have two main procedures: updating empirical distributions and constructing CBs. Their dependence on $V$ is trackable because each step’s contribution can be isolated. In contrast, the EL-LUCB method directly builds upper and lower bounds at each iteration rather than separately constructing CBs, making the role of $V$ less transparent. We illustrate $V$’s impact through various examples.

First, we run the EL-LUCB (EL) algorithm for previous cases. Interestingly, it remains robust across both supports, showing consistent performance despite changes in the distribution’s support. It outperforms both Structured-LUCB and Non-structured-LUCB:
 \begin{figure}[H]
    \begin{subfigure}{}
        \includegraphics[width=0.49\textwidth]{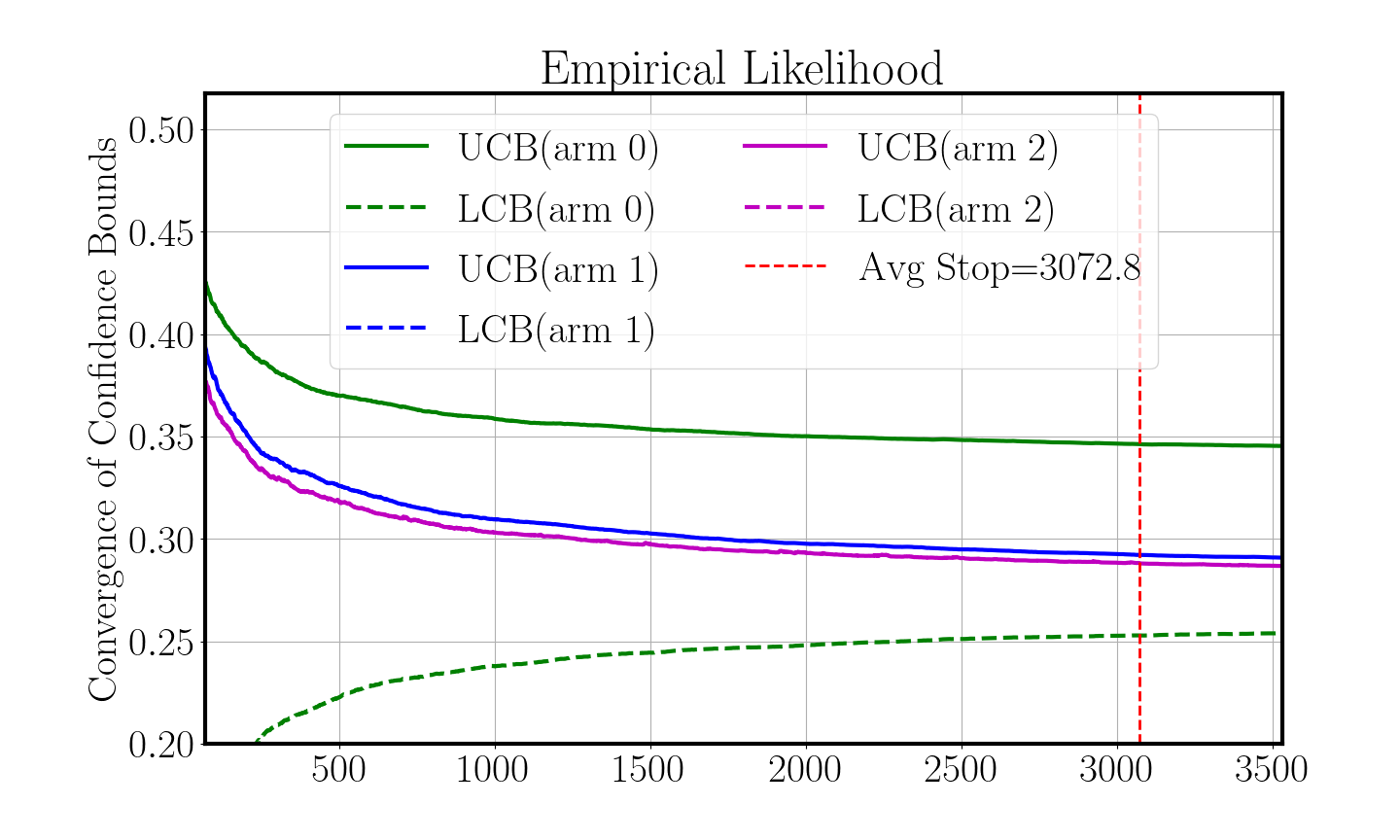}
        \label{EL_stwin}
    \end{subfigure}
\begin{subfigure}{}
 \includegraphics[width=0.49\textwidth]{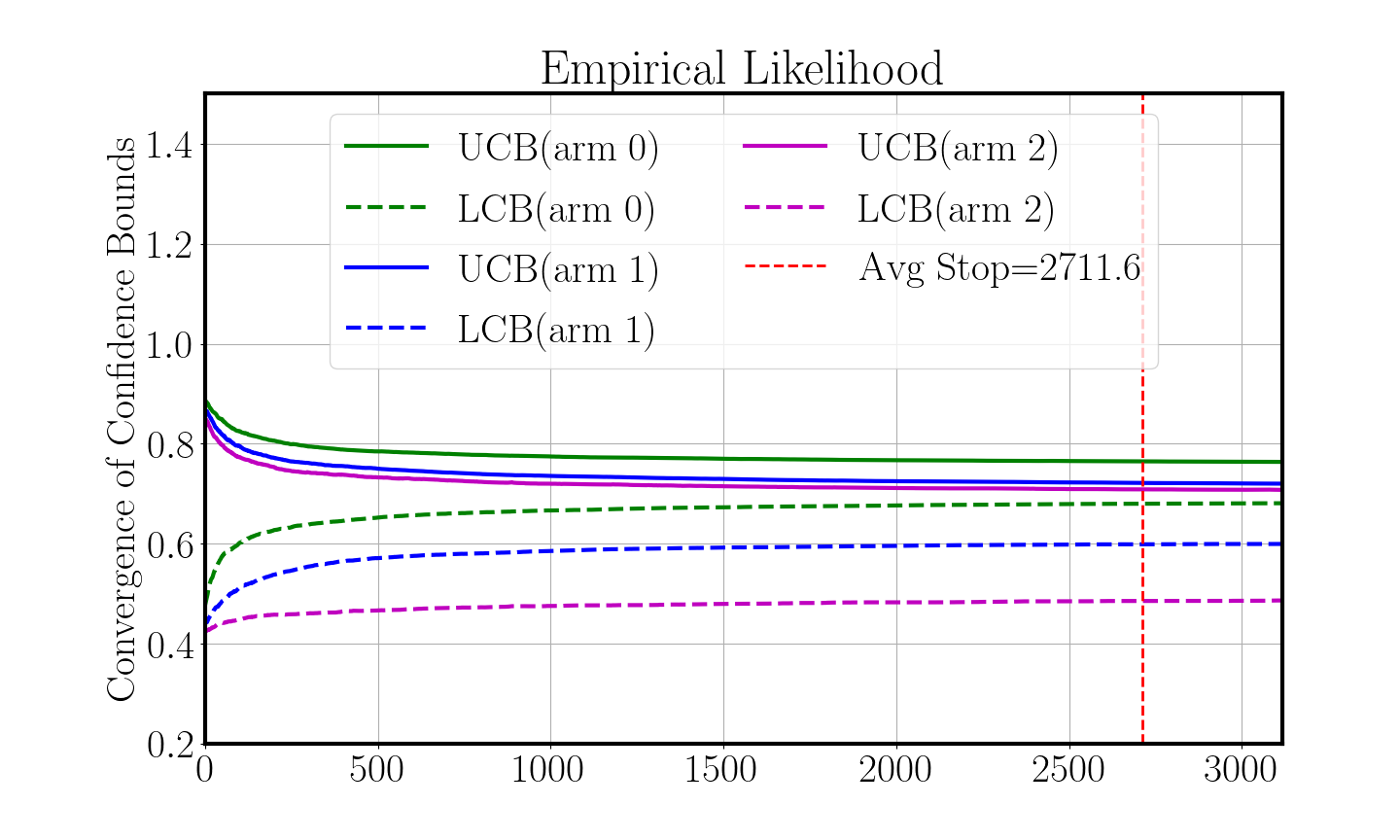}
 \end{subfigure}
    \caption{Comparing the stopping times of EL-LUCB algorithm on $V^{\text{test}1}$ (above) and $V^{\text{test}2}$ (below). }
    \label{EL_nonstwin}
\end{figure}
In these examples, EL-LUCB works on a same range $\text{Range}(V)=\max(V)-\min(V)$ on the support and seems  the condition $\KL(\cdot,\cdot)\leq \epsilon$ as we can see does not directly by each component of $V$. 
Consequently, we propose two cases where the distinguishability of two arms remains close and we  show how the performance is affected by the range of $V$. 
We consider
\[
P^{\text{test2}} =
\begin{bmatrix}
0.142 & 0.311 & 0.153 & 0.391 \\
0.386 & 0.114 & 0.154 & 0.344
\end{bmatrix},
\]
\[
V^{\text{test}3} = 
\begin{bmatrix}
0.144, 0.152, 0.505, 0.984
\end{bmatrix},
\]
\[
V^{\text{test}4}= 
\begin{bmatrix}
0.573, 0.518,0.409, 0.505
\end{bmatrix},
\]
where the  range of $V$ changes from $\Delta V^{\text{test}3} = 0.84$ to $\Delta V^{\text{test}4} = 0.164$. 

We tailored these two cases so that the expected rewards of two arms remain close, yielding  $\mtbE_{P[0]}[V^{\text{test}3}]-\mtbE_{P[1]}[V^{\text{test}3}] = 0.04$ and $\mtbE_{P[0]}[V^{\text{test}4}]-\mtbE_{P[1]}[V^{\text{test}4}] = 0.014$. 
\begin{figure}[H]
    \centering
    \begin{subfigure}{}
\includegraphics[width=0.49\textwidth]{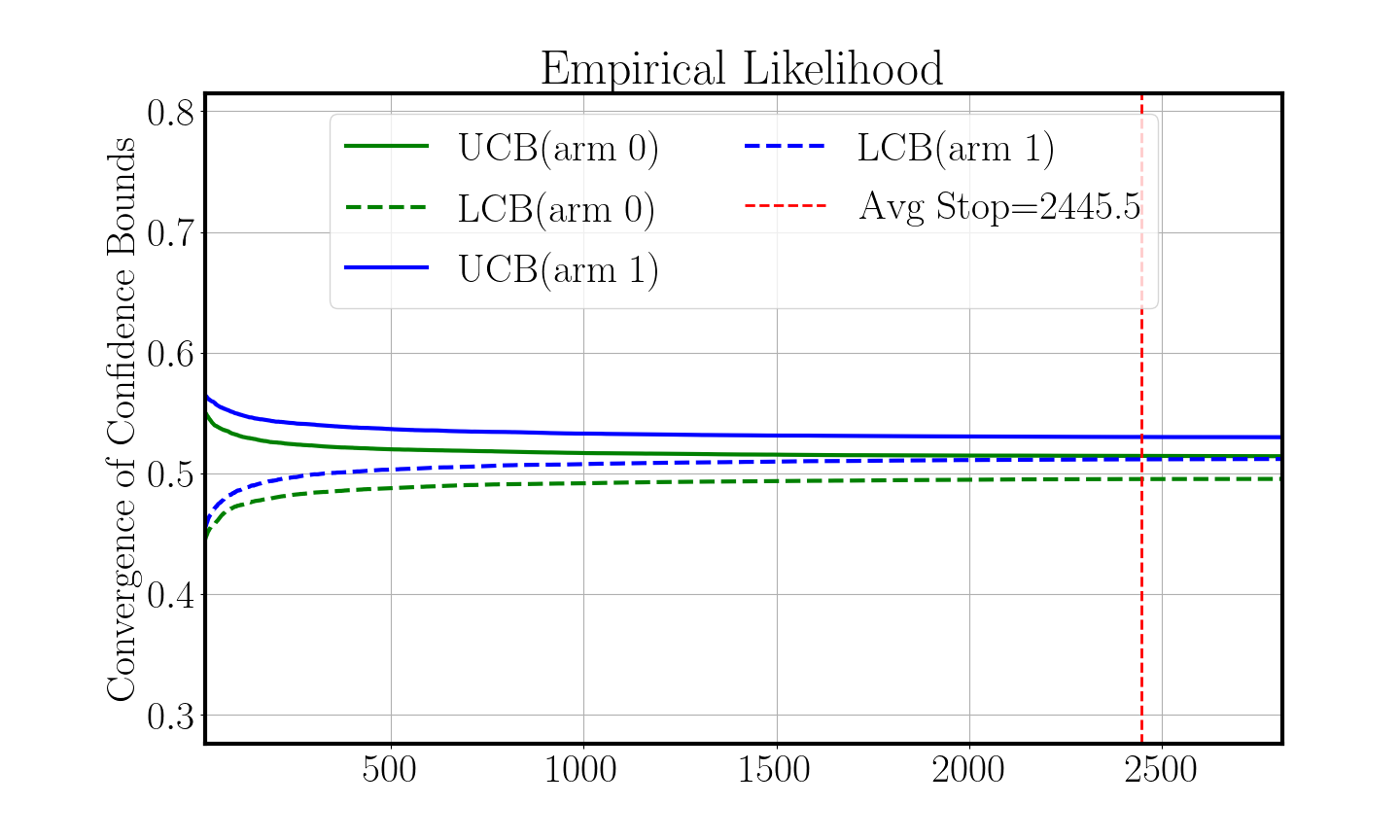}
    \end{subfigure}
    \begin{subfigure}{}
        \includegraphics[width=0.5\textwidth]{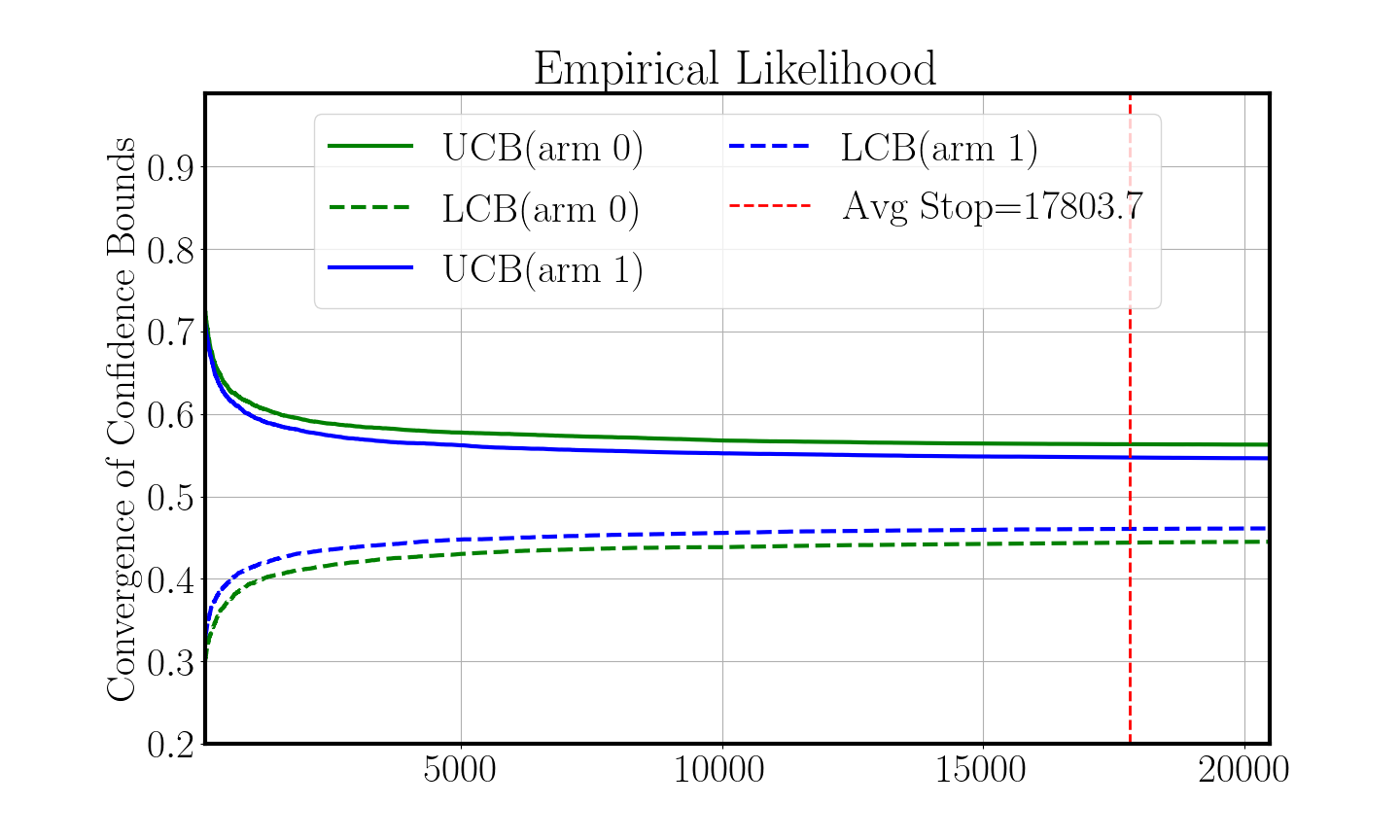}
    \end{subfigure}
    \caption{Comparing the stopping times of EL-LUCB algorithm on $V^\text{test3}$ with low range (above) and with $V^\text{test4}$ high range (below)}
\label{ex1:smallV}
\end{figure}
 It appears that EL-LUCB    's stopping time is lower than the previous approaches  in both cases, at the price of a somewhat higher  computational complexity for solving several Lagrangian problems.
\section{Conclusion}
We presented and compared two multi-armed bandit (MAB) frameworks to tackle the choice of a transition in a finite horizon Markov decision process with known future values. 
Experimental results compared stopping times and the impact of leveraging distributional support across various scenarios, illustrating that incorporating structure -- when available -- may significantly improve performance and decision-making efficiency. The quantitative amplitude of this gain remains to be better understood from a theoretical point of view.  

\nocite{langley00}

\bibliography{main}

\begin{thebibliography}{26}
\providecommand{\natexlab}[1]{#1}
\providecommand{\url}[1]{\texttt{#1}}
\expandafter\ifx\csname urlstyle\endcsname\relax
  \providecommand{\doi}[1]{doi: #1}\else
  \providecommand{\doi}{doi: \begingroup \urlstyle{rm}\Url}\fi

\bibitem[Abbasi-Yadkori et~al.(2011)Abbasi-Yadkori, P{\'a}l, and
  Szepesv{\'a}ri]{abbasi2011improved}
Abbasi-Yadkori, Y., P{\'a}l, D., and Szepesv{\'a}ri, C.
\newblock Improved algorithms for linear stochastic bandits.
\newblock \emph{Advances in neural information processing systems}, 24, 2011.

\bibitem[Agrawal et~al.(2020)Agrawal, Juneja, and Glynn]{agrawal2020optimal}
Agrawal, S., Juneja, S., and Glynn, P.
\newblock Optimal $\delta$-correct best-arm selection for heavy-tailed
  distributions.
\newblock In \emph{Algorithmic Learning Theory}, pp.\  61--110. PMLR, 2020.

\bibitem[Audibert \& Bubeck(2010)Audibert and Bubeck]{audibert2010best}
Audibert, J.-Y. and Bubeck, S.
\newblock Best arm identification in multi-armed bandits.
\newblock In \emph{COLT-23th Conference on learning theory-2010}, pp.\  13--p,
  2010.

\bibitem[Bertsekas(2005)]{bertsekas2005dp}
Bertsekas, D.~P.
\newblock \emph{Dynamic Programming and Optimal Control}.
\newblock Athena Scientific, Belmont, MA, 3rd edition, 2005.
\newblock ISBN 978-1886529267.

\bibitem[Bubeck et~al.(2012)Bubeck, Cesa-Bianchi, et~al.]{bubeck2012regret}
Bubeck, S., Cesa-Bianchi, N., et~al.
\newblock Regret analysis of stochastic and nonstochastic multi-armed bandit
  problems.
\newblock \emph{Foundations and Trends{\textregistered} in Machine Learning},
  5\penalty0 (1):\penalty0 1--122, 2012.

\bibitem[Capp{\'e} et~al.(2013)Capp{\'e}, Garivier, Maillard, Munos, and
  Stoltz]{cappe2013kullback}
Capp{\'e}, O., Garivier, A., Maillard, O.-A., Munos, R., and Stoltz, G.
\newblock Kullback-leibler upper confidence bounds for optimal sequential
  allocation.
\newblock \emph{The Annals of Statistics}, pp.\  1516--1541, 2013.

\bibitem[Chen et~al.(2024)Chen, Hu, Zhao, Wang, and Qiao]{car}
Chen, S., Hu, X., Zhao, J., Wang, R., and Qiao, M.
\newblock A review of decision-making and planning for autonomous vehicles in
  intersection environments.
\newblock \emph{World Electric Vehicle Journal}, 15\penalty0 (3), 2024.
\newblock ISSN 2032-6653.
\newblock \doi{10.3390/wevj15030099}.
\newblock URL \url{https://www.mdpi.com/2032-6653/15/3/99}.

\bibitem[Filippi et~al.(2010)Filippi, Cappé, and Garivier]{SaraF2010}
Filippi, S., Cappé, O., and Garivier, A.
\newblock Optimism in reinforcement learning and kullback-leibler divergence.
\newblock In \emph{2010 48th Annual Allerton Conference on Communication,
  Control, and Computing (Allerton)}, pp.\  115--122, 2010.
\newblock \doi{10.1109/ALLERTON.2010.5706896}.

\bibitem[Gabillon et~al.(2012)Gabillon, Ghavamzadeh, and
  Lazaric]{gabillon2012best}
Gabillon, V., Ghavamzadeh, M., and Lazaric, A.
\newblock Best arm identification: A unified approach to fixed budget and fixed
  confidence.
\newblock \emph{Advances in Neural Information Processing Systems}, 25, 2012.

\bibitem[Garivier \& Kaufmann(2016)Garivier and Kaufmann]{Garivier16}
Garivier, A. and Kaufmann, E.
\newblock Optimal best arm identification with fixed confidence.
\newblock In Feldman, V., Rakhlin, A., and Shamir, O. (eds.), \emph{29th Annual
  Conference on Learning Theory}, volume~49 of \emph{Proceedings of Machine
  Learning Research}, pp.\  998--1027, Columbia University, New York, New York,
  USA, June 2016. PMLR.
\newblock URL \url{https://proceedings.mlr.press/v49/garivier16a.html}.

\bibitem[Heidrich-Meisner \& Igel(2009)Heidrich-Meisner and
  Igel]{Heidrich2009Bernstein}
Heidrich-Meisner, V. and Igel, C.
\newblock Hoeffding and bernstein races for selecting policies in evolutionary
  direct policy search.
\newblock In \emph{Proceedings of the 26th Annual International Conference on
  Machine Learning}, ICML '09, pp.\  401–408, New York, NY, USA, 2009.
  Association for Computing Machinery.
\newblock ISBN 9781605585161.
\newblock \doi{10.1145/1553374.1553426}.
\newblock URL \url{https://doi.org/10.1145/1553374.1553426}.

\bibitem[Honda \& Takemura(2010)Honda and Takemura]{Honda2010AnAO}
Honda, J. and Takemura, A.
\newblock An asymptotically optimal bandit algorithm for bounded support
  models.
\newblock In \emph{Annual Conference Computational Learning Theory}, 2010.
\newblock URL \url{https://api.semanticscholar.org/CorpusID:120162138}.

\bibitem[Jourdan et~al.(2022)Jourdan, Degenne, Baudry, de~Heide, and
  Kaufmann]{jourdan2022top}
Jourdan, M., Degenne, R., Baudry, D., de~Heide, R., and Kaufmann, E.
\newblock Top two algorithms revisited.
\newblock \emph{Advances in Neural Information Processing Systems},
  35:\penalty0 26791--26803, 2022.

\bibitem[Kalyanakrishnan et~al.(2012)Kalyanakrishnan, Tewari, Auer, and
  Stone]{Kalyanakrishnan2012LUCB}
Kalyanakrishnan, S., Tewari, A., Auer, P., and Stone, P.
\newblock Pac subset selection in stochastic multi-armed bandits.
\newblock In \emph{International Conference on Machine Learning}, 2012.
\newblock URL \url{https://api.semanticscholar.org/CorpusID:1635758}.

\bibitem[Kaufmann \& Kalyanakrishnan(2013)Kaufmann and
  Kalyanakrishnan]{kaufmann2013KLLUCB}
Kaufmann, E. and Kalyanakrishnan, S.
\newblock Information complexity in bandit subset selection.
\newblock In \emph{Conference on Learning Theory}, pp.\  228--251. PMLR, 2013.

\bibitem[Lopes \& Lopes(2022)Lopes and Lopes]{DDAsurvey}
Lopes, J.~C. and Lopes, R.~P.
\newblock A review of dynamic difficulty adjustment methods for serious
  games.
\newblock In Pereira, A.~I., Ko{\v{s}}ir, A., Fernandes, F.~P., Pacheco, M.~F.,
  Teixeira, J.~P., and Lopes, R.~P. (eds.), \emph{Optimization, Learning
  Algorithms and Applications}, pp.\  144--159, Cham, 2022. Springer
  International Publishing.
\newblock ISBN 978-3-031-23236-7.

\bibitem[Lu et~al.(2010)Lu, P{\'a}l, and P{\'a}l]{lu2010contextual}
Lu, T., P{\'a}l, D., and P{\'a}l, M.
\newblock Contextual multi-armed bandits.
\newblock In \emph{Proceedings of the Thirteenth international conference on
  Artificial Intelligence and Statistics}, pp.\  485--492. JMLR Workshop and
  Conference Proceedings, 2010.

\bibitem[Mnih et~al.(2008)Mnih, Szepesv\'{a}ri, and
  Audibert]{Minh2008Bernstein}
Mnih, V., Szepesv\'{a}ri, C., and Audibert, J.-Y.
\newblock Empirical bernstein stopping.
\newblock In \emph{Proceedings of the 25th International Conference on Machine
  Learning}, ICML '08, pp.\  672–679, New York, NY, USA, 2008. Association
  for Computing Machinery.
\newblock ISBN 9781605582054.
\newblock \doi{10.1145/1390156.1390241}.
\newblock URL \url{https://doi.org/10.1145/1390156.1390241}.

\bibitem[Moerland et~al.(2023{\natexlab{a}})Moerland, Broekens, Plaat, and
  Jonker]{RL-survey}
Moerland, T.~M., Broekens, J., Plaat, A., and Jonker, C.~M.
\newblock Model-based reinforcement learning: A survey.
\newblock \emph{Foundations and Trends® in Machine Learning}, 16\penalty0
  (1):\penalty0 1--118, 2023{\natexlab{a}}.
\newblock ISSN 1935-8237.
\newblock \doi{10.1561/2200000086}.
\newblock URL \url{http://dx.doi.org/10.1561/2200000086}.

\bibitem[Moerland et~al.(2023{\natexlab{b}})Moerland, Broekens, Plaat, Jonker,
  et~al.]{moerland2023model}
Moerland, T.~M., Broekens, J., Plaat, A., Jonker, C.~M., et~al.
\newblock Model-based reinforcement learning: A survey.
\newblock \emph{Foundations and Trends{\textregistered} in Machine Learning},
  16\penalty0 (1):\penalty0 1--118, 2023{\natexlab{b}}.

\bibitem[Neu et~al.(2024)Neu, Olkhovskaya, and Vakili]{Vakili2024}
Neu, G., Olkhovskaya, J., and Vakili, S.
\newblock Adversarial contextual bandits go kernelized.
\newblock In Vernade, C. and Hsu, D. (eds.), \emph{Proceedings of The 35th
  International Conference on Algorithmic Learning Theory}, volume 237 of
  \emph{Proceedings of Machine Learning Research}, pp.\  907--929. PMLR, 25--28
  Feb 2024.
\newblock URL \url{https://proceedings.mlr.press/v237/neu24a.html}.

\bibitem[Puterman(1994)]{puterman1994mdp}
Puterman, M.~L.
\newblock \emph{Markov Decision Processes: Discrete Stochastic Dynamic
  Programming}.
\newblock John Wiley \& Sons, New York, 1994.
\newblock ISBN 978-0471727828.

\bibitem[Russo(2016)]{russo2016simple}
Russo, D.
\newblock Simple bayesian algorithms for best arm identification.
\newblock In \emph{Conference on Learning Theory}, pp.\  1417--1418. PMLR,
  2016.

\bibitem[Saber \& Maillard(2024)Saber and Maillard]{saber2024bandits}
Saber, H. and Maillard, O.-A.
\newblock Bandits with multimodal structure.
\newblock In \emph{Reinforcement Learning Conference}, volume~1, pp.\ ~39,
  2024.

\bibitem[Sutton(2018)]{sutton2018reinforcement}
Sutton, R.~S.
\newblock Reinforcement learning: An introduction.
\newblock \emph{A Bradford Book}, 2018.

\bibitem[You et~al.(2023)You, Qin, Wang, and Yang]{you2023information}
You, W., Qin, C., Wang, Z., and Yang, S.
\newblock Information-directed selection for top-two algorithms.
\newblock In \emph{The Thirty Sixth Annual Conference on Learning Theory}, pp.\
   2850--2851. PMLR, 2023.

\end{thebibliography}
\bibliographystyle{icml2025} 


\end{document}